\pdfoutput=1 
\PassOptionsToPackage{dvipsnames,table}{xcolor}
\documentclass[letterpaper, 10 pt, conference]{ieeeconf}  
\usepackage{textcomp}
\usepackage{subcaption}
\usepackage{amsfonts,amssymb} 
\usepackage{booktabs}
\usepackage{stfloats}
\usepackage{url}
\usepackage{verbatim}
\usepackage{cite}
\usepackage[version=4]{mhchem}
\usepackage{graphicx} 
\usepackage{epsfig} 
\usepackage{mathptmx} 
\usepackage{times} 
\usepackage{amsmath} 
\usepackage[T1]{fontenc}
\usepackage{tabularx}

\usepackage{multirow}
\usepackage{makecell}
\usepackage{diagbox} 
\usepackage{bm}
\usepackage{float}
\usepackage{caption}
\usepackage{cuted}
\usepackage{circledsteps}
\usepackage{siunitx}
\usepackage{pifont}    
\usepackage{array}

\usepackage{xcolor}

\usepackage{hyperref}
\IEEEoverridecommandlockouts                              
\title{\LARGE \bf
Exo-ViHa: A Cross-Platform Exoskeleton System with Visual and Haptic Feedback for Efficient Dexterous Skill Learning
}

\author{
    Xintao Chao\textsuperscript{*1}, Shilong Mu\textsuperscript{*†1}, Yushan Liu\textsuperscript{1}, Shoujie Li\textsuperscript{1} , Chuqiao Lyu\textsuperscript{†1}, \\
     Xiao-Ping Zhang\textsuperscript{1}, \textit{Fellow, IEEE}, Wenbo Ding\textsuperscript{†1}
}

\begin{document}
\maketitle
\begingroup
\renewcommand\thefootnote{}

\footnotetext{*These authors contributed equally to this work.}
\footnotetext{†Corresponding authors.}
\footnotetext{\textsuperscript{1}Shenzhen International Graduate School, Tsinghua University}
\endgroup
\thispagestyle{empty}
\pagestyle{empty}

\begin{strip}
  \vspace{-20mm}
  \centering
  \includegraphics[width=1.0\textwidth]{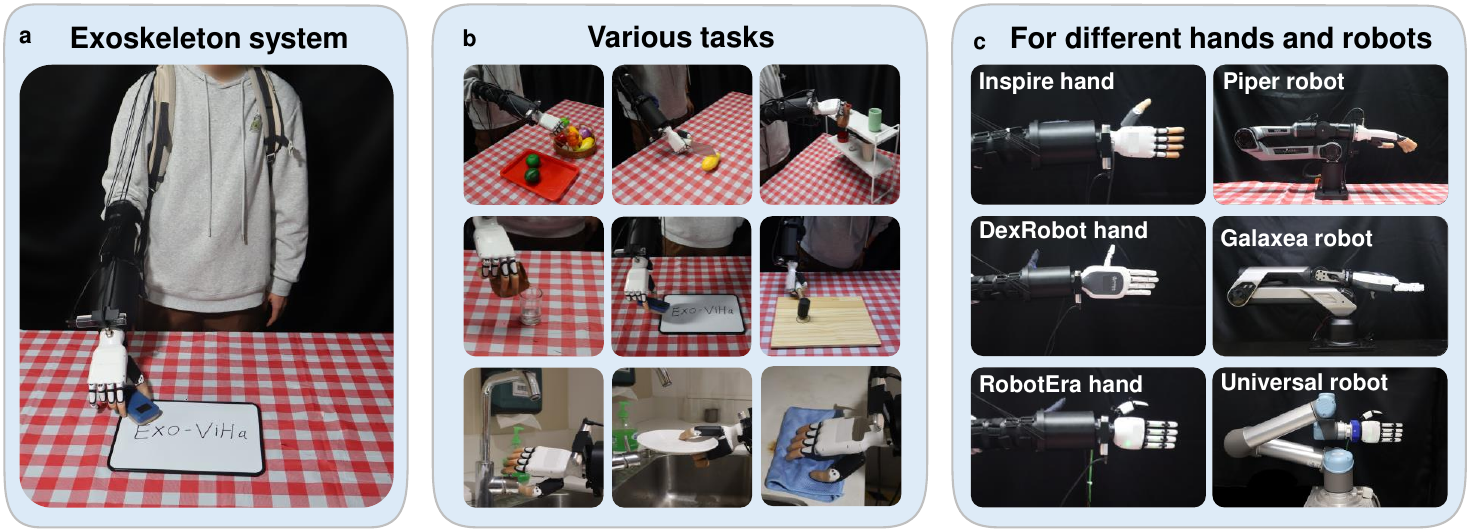}
  \captionof{figure}{\textbf{Overview of the Exo-ViHa system.}  (a) Demonstration of the exoskeleton-based data collection system.  (b) Data collection of diverse tasks in various scenarios.  (c) Compatibility of the Exo-ViHa system with different dexterous robotic hands and arms.}
  \label{fig1}
\end{strip}

\begin{abstract}
Imitation learning has emerged as a powerful paradigm for robot skills learning. However, traditional data collection systems for dexterous manipulation face challenges, including a lack of balance between acquisition efficiency, consistency, and accuracy. To address these issues, we introduce Exo-ViHa, an innovative 3D-printed exoskeleton system that enables users to collect data from a first-person perspective while providing real-time haptic feedback. This system combines a 3D-printed modular structure with a slam camera, a motion capture glove, and a wrist-mounted camera. Various dexterous hands can be installed at the end, enabling it to simultaneously collect the posture of the end effector, hand movements, and visual data. By leveraging the first-person perspective and direct interaction, the exoskeleton enhances the task realism and haptic feedback, improving the consistency between demonstrations and actual robot deployments. In addition, it has cross-platform compatibility with various robotic arms and dexterous hands. Experiments show that the system can significantly improve the success rate and efficiency of data collection for dexterous manipulation tasks. Webpage: \url{https://exo-viha2025.github.io/}.

\end{abstract}

\section{Introduction}

Dexterous manipulation remains a key challenge in the field of robotics, particularly for tasks requiring fine control and high degree-of-freedom coordination, such as pen spinning \cite{nakatani2023dynamic}, ball throwing\cite{munn2024whole}, and folding clothes\cite{munn2024whole}. To enable robots to learn these complex tasks, imitation learning provides an effective solution by allowing robots to learn various skills from human demonstration\cite{chi2023diffusion,ze20243d}. However, the effectiveness of imitation learning largely depends on high-quality, large-scale demonstration data, which imposes higher requirements on data collection systems.

Traditional data collection methods often rely on teleoperation systems, such as virtual reality (VR) devices \cite{arunachalam2023dexterous, arunachalam2023holo},\cite{cheng2024open}, handheld controllers \cite{rakita2019remote, Khadir2019TeleoperatorIW}, or vision-based teleoperation \cite{li2022dexterous, qin2023anyteleop},\cite{Handa2019DexPilotVT}. These approaches improve the flexibility of teleoperate to some extent, allowing humans to intuitively control robots and collect corresponding manipulation data. However, they typically suffer from limitations such as the lack of low-latency first-person visual feedback, the absence of haptic feedback, and inconsistencies between the data collection and deployment phases. These limitations may lead to inefficient data collection or degraded model performance during real-world deployment, thus reducing the generalizability of imitation learning.

To address these constraints, we introduce a novel exoskeleton device designed to efficiently collect data for dexterous hand imitation learning tasks. The proposed exoskeleton is a 3D-printed component designed to accommodate various dexterous hands. During data collection, the user first mounts a  Intel RealSense tracking camera T265 and a wrist-mounted camera onto the exoskeleton’s camera base. Then, the user wears a motion capture glove and attaches the exoskeleton to the forearm to initiate data collection. The system collects multiple types of data, including end-effector pose data from the T265 camera, motion data from the dexterous hand, and frame data from the cameras. We employ the Action Chunking Transformer (ACT) \cite{lee2023dexterous} to train the imitation learning algorithm. The output end-effector pose and dexterous hand motion data from the model are then used to control the robotic arm and dexterous hand for movement and manipulation.

\begin{table*}[ht]
\centering
\caption{Comparison of Data Collection Systems for Dexterous Manipulation}
\label{tab:comparison}
\renewcommand{\arraystretch}{1.2}
\setlength{\tabcolsep}{4pt}
\begin{tabular}{lcccccccl}
\toprule
\textbf{System} & \textbf{End-Effector} & \textbf{Visual} & \textbf{Haptic} & \textbf{VC} & \textbf{BC} & \textbf{Data Type} & \textbf{Cross-Platform} & \textbf{Cost} \\
\midrule
DexCap \cite{dexcap2024} & Dexterous Hand  & FP  & 
$\star\star\star\star\star$ & 
$\star\star$ & 
$\star\star$ & Off-line & \ding{55} & Low \\
\rowcolor{gray!20} UMI\cite{umi2024} & Gripper & FP & 
$\star\star\star$ & 
$\star\star\star\star\star$ & 
$\star\star\star\star$ & Off-line & \ding{55} & Low \\
ForceMimic \cite{forcemimic2024}& Gripper & FP  & 
$\star\star\star$ & 
$\star\star\star\star\star$ & 
$\star\star\star\star$ & Off-line & \ding{55} & Medium \\
\rowcolor{gray!20} Fast-UMI\cite{wu2024fast} & Gripper & FP & 
$\star\star\star$ & 
$\star\star\star\star\star$ & 
$\star\star\star\star$ & Off-line & \ding{55} & Low \\
Mobile ALOHA\cite{mobilealoha2024} & Gripper & TP  & 
$\star\star$ & 
$\star\star\star\star\star$ & 
$\star\star\star\star\star$ & On-line & \ding{55} & High \\
\rowcolor{gray!20} Feelit\cite{yu2024feelit} & Gripper & TP & 
$\star\star\star\star$ & 
$\star\star\star\star\star$ & 
$\star\star\star\star\star$ & On-line & \ding{55} & High \\
AnyTeleop\cite{qin2023anyteleop} & Dexterous Hand & VR  & 
$\star$ & 
$\star\star\star\star\star$ & 
$\star\star\star\star\star$ & On-line & \ding{51} & High \\
\rowcolor{gray!20} Open-TeleVision\cite{cheng2024open} & Dexterous Hand & VR & 
$\star$ & 
$\star\star\star\star\star$ & 
$\star\star\star\star\star$ & On-line & \ding{51} & High \\
Bunny-VisionPro\cite{ding2024bunny} & Dexterous Hand & VR  & 
$\star\star\star$ & 
$\star\star\star\star\star$ & 
$\star\star\star\star\star$ & On-line & \ding{51} & High \\
\rowcolor{gray!20} ACE\cite{ace2024} & Hand/Gripper & TP  & 
$\star$ & 
$\star\star\star\star\star$ & 
$\star\star\star\star\star$ & On-line & \ding{51} & High \\
HIRO\cite{wei2024wearable} & Dexterous Hand & FP & 
$\star\star\star\star$ & 
$\star\star$ & 
$\star\star$ & On-line & \ding{55} & Low \\
\hline
\rowcolor{gray!20} \textbf{Our System} & \textbf{Dexterous Hand} & \textbf{FP} & 
\textbf{$\star\star\star$} & 
\textbf{$\star\star\star\star\star$} & 
\textbf{$\star\star\star\star$} & \textbf{Off-line} & \textbf{\ding{51}} & \textbf{Low} \\
\bottomrule
\end{tabular}

\vspace{4pt}
\footnotesize
FP: First Person, VR: Virtual Reality, TP: Third Person, 
VC: Visual Consistency (rated $\star$ to $\star\star\star\star\star$), \\
BC: Body Consistency (rated $\star$ to $\star\star\star$). 
Higher star count indicates better performance.
\end{table*}
To properly evaluate our contributions, we evaluate our system against existing data collection solutions, considering several critical dimensions as illustrated in Table \ref{tab:comparison}.

The advantages of our system are highlighted through comparisons with existing data collection solutions:

\begin{itemize}
    \item \textbf{First-Person visual feedback}: Our exoskeleton offers true first-person visual feedback. This direct engagement allows users to operate in their natural field of view, enhancing task realism and operational accuracy.
    
    \item \textbf{Haptic feedback}: Since the exoskeleton makes direct contact with the user's arm, users can distinctly perceive tactile sensations and force feedback while operating the objects, which  is crucial for tasks that require force adjustment during the interaction with multiple objects.
    
    \item \textbf{Visual consistency}: By placing identical camera bases on both the exoskeleton and the robotic arm, cameras can collect sufficient visual information while ensuring a consistent visual perspective during both data collection and deployment phases. 
    
    \item \textbf{Body consistency}: Our system directly collects data from the dexterous hand when performing tasks. This contrasts with other platforms where data might be indirectly mapped or transformed, ensuring that our deployment results are more reliable and closely aligned with the demonstration conditions.
    
\end{itemize}


\section{Related Work}
\label{sec:related_work}

\subsection{Imitation Learning }
Imitation Learning (IL) has gained significant traction in recent years, primarily due to its ability to mimic expert behavior in complex environments, providing a pathway to efficient learning without exhaustive trial-and-error exploration. This paradigm demonstrates versatility across various domains, including robotics, autonomous driving, and game playing, showcasing its ability to generalize expert knowledge effectively\cite{cheng2024pluto,zhang2022improve}.
The IL method can have input from different sources of demonstrations, such as human demonstrations\cite{xiong2024adaptive,mendonca2023structured}, real-Robot expert data\cite{mobilealoha2024,lee2023dexterous,brohan2022rt}, and simulation date\cite{hathaway2023imitation,liu2022interactive,guo2024lasil}, thus enhancing its practicality in real production.
Several algorithms have been developed to improve the performance of IL systems. Among them, the Differential Privacy (DP)\cite{chi2023diffusion} algorithm and the ACT algorithm stand out, which has proven effective in tasks that require high temporal precision and accurate control. 

\subsection{Data Collection Systems}
In recent years, several data collection systems have been developed for robotic applications.  One common method is teleoperation\cite{qin2023anyteleop,chi2023diffusion,caeiro2021systematic}, where robots are controlled manually from a distance.  Another approach involves the use of haptic devices, which provide tactile feedback, allowing the operator to feel the robot's interactions with the environment.  Virtual reality systems also offer an immersive experience for data collection\cite{arunachalam2023dexterous, arunachalam2023holo},\cite{cheng2024open,ding2024bunny}, enabling users to control robots in a simulated environment. 

For imitation learning, it is crucial to ensure that input-output data remain consistent to achieve effective learning. High-quality data collection should prioritize precision, which often requires tactile feedback and first-person perspectives to accurately capture the robot’s interactions. However, many existing systems lack these features, which limits the effectiveness of the collected data for training sophisticated models.

\section{Exo-ViHa System Architecture}
\subsection{Hardware Design}
\begin{figure*}[thpb]
  \centering
  \includegraphics[width=1.0\linewidth]{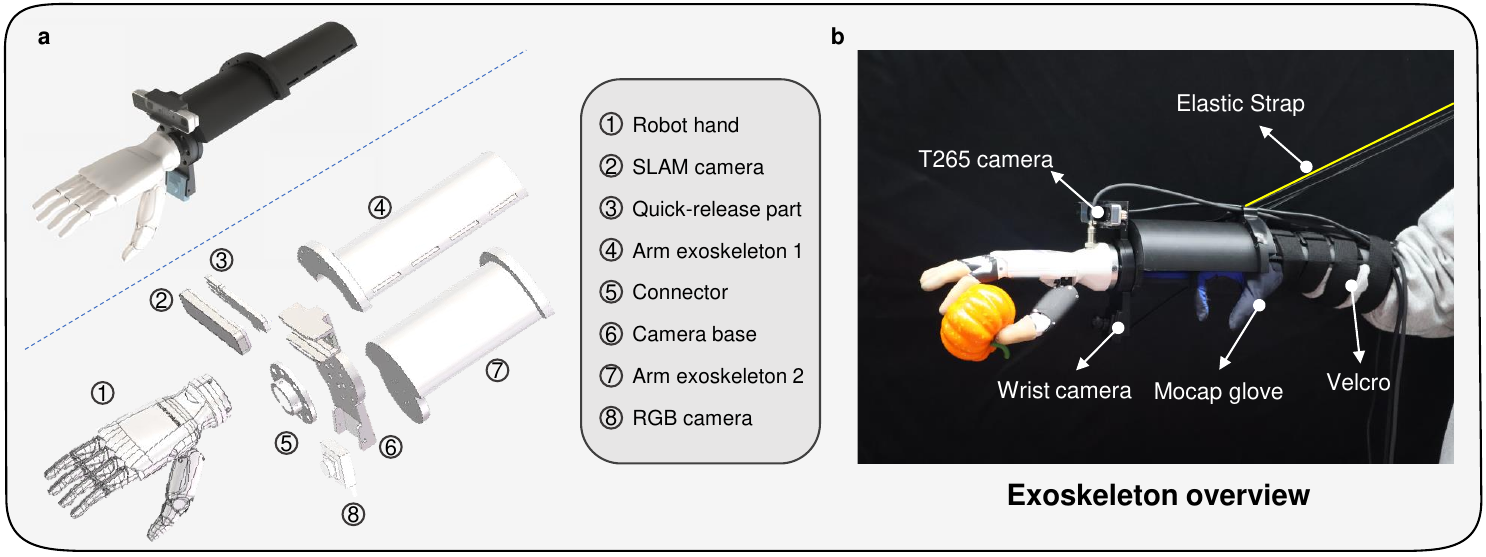}
  \caption{\textbf{Details of the exoskeleton system.} (a) The components of the exoskeleton system. (b) Exoskeleton wearing demonstration.
}\label{fig3}
\end{figure*}
The design of the exoskeleton system is illustrated in Fig.~\ref{fig3}. The entire exoskeleton is constructed using 3D printing technology with Polylactic Acid (PLA) material, ensuring a balance of durability and lightweight structure. The overall weight of the system is about 425 grams. Upon first use, users can assemble all components within a few minutes. The precise 3D printing process ensures that all components fit seamlessly and securely, providing both functionality and comfort for the user.

The forearm exoskeleton 1 is secured to the user's hand using velcro straps, while forearm exoskeleton 2 provides cover and finger movement space for the user's hand during data collection, ensuring that the external cameras can only view the dexterous hand and not the user's hand. Additionally, the camera base component includes slots for the T265 camera and the wrist camera below. To facilitate quick installation and removal of the cameras, the slots feature a rapid insertion clip design, allowing the T265 camera to be easily attached and detached using a compatible quick-connect adapter. 

Moreover, the system is designed to accommodate a wide range of dexterous hands by 3D printing specific connectors tailored for each particular hand or two-finger gripper. The entire exoskeleton system weighs approximately 425 grams; however, when a dexterous hand is attached, the total weight increases to around 1100 grams.To address this, elastic bands are utilized to provide a degree of gravitational compensation, thereby reducing the burden on the user's arm.

\subsection{System Design}
\begin{figure*}[thpb]
  \centering
  \includegraphics[width=1.0\linewidth]{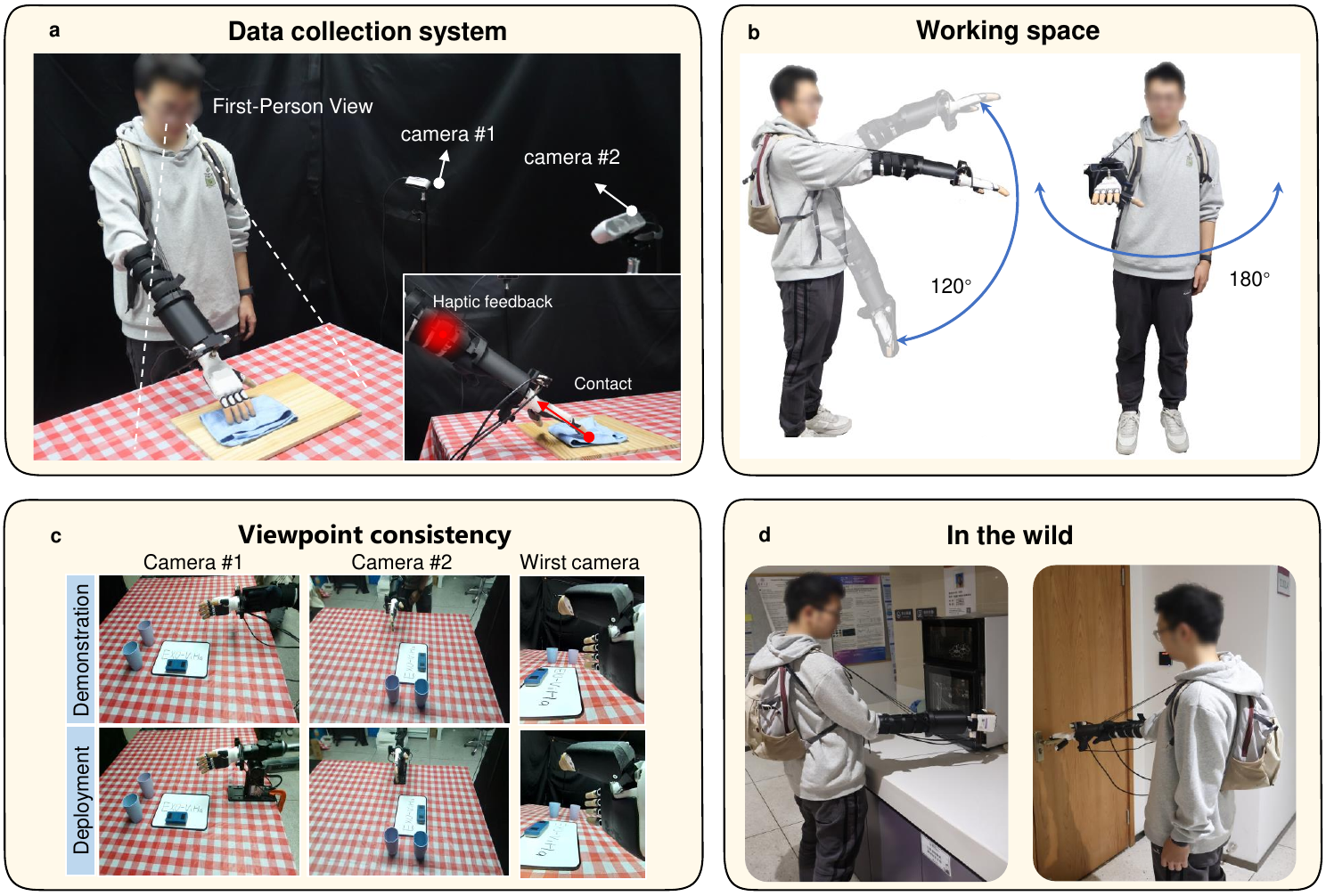}
  \caption{\textbf{Details of the Exo-ViHa system.} (a) Demonstration of the exoskeleton used for data collection and schematic diagram of the first-person perspective and haptic feedback. (b) Illustration of the working space range. (c) The visual information captured by the external cameras and the wrist camera is basically consistent during both the data collection and deployment phases. (d) The power source and microcomputer are placed into a backpack, allowing the system to be used for data collection in outdoor environments.
}\label{fig2}
\end{figure*}
The operational demonstration of the entire Exo-ViHa system is illustrated in Fig.~\ref{fig2}. During the data collection phase, in addition to the SLAM camera and wrist camera mounted on the exoskeleton, we also employed two external cameras positioned at different angles to comprehensively capture environmental information. As shown in Fig.~\ref{fig2} (a), during data collection, experimenters can access first-person visual information, allowing them to observe the current workspace through their own eyes and control the position and posture of the dexterous hand, facilitating a more direct acquisition of spatial information.

In addition, when performing tasks with the exoskeleton and dexterous hand, we experience a noticeable resistance through the dexterous hand to our forearm when it makes contact with objects, even if there are no haptic sensors to capture specific contact pressure. This force feedback enables us to perceive the interactions and friction between objects effectively during multi-contact tasks, allowing for better control of forearm positioning and the use of the dexterous hand, thereby enhancing the quality of the data demonstration.

While wearing the entire exoskeleton, our arms maintain a significant range of motion despite the presence of elastic bands used for gravitational compensation. As shown in Fig.~\ref{fig2} (b), wearers can easily rotate their arms 180° from side to side. The vertical motion is limited to approximately 120° primarily due to the length of the connecting cables. However, this range of motion is effectively unrestricted for performing basic tasks.

Another advantage of the system lies in the consistency of perspective. During data collection, the arm wearing the exoskeleton is largely obscured, making it difficult to view it directly. Consequently, external cameras can capture information about the positioning and tasks of the dexterous hand more effectively. 
\begin{figure}[htbp]
    \centering
    \includegraphics[width=0.45\textwidth]{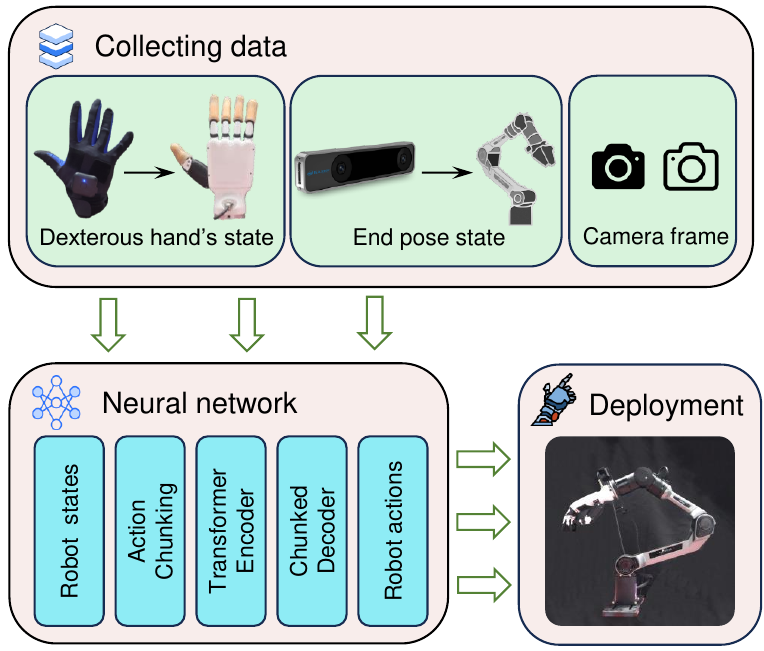}
    \caption{\textbf{Workflow of data collection and deployment of Exo-ViHa.} The collected data include end-effector states, 6D end-effector pose and camera frames. }
    \label{fig4}
\end{figure}
Additionally, since our camera base is mounted on both the exoskeleton and the robotic arm, as illustrated in Fig.~\ref{fig2}  (c), the information obtained by the wrist camera is nearly identical as long as the end-effector's pose remains consistent during both collection and deployment. This alignment allows the model's outputs to closely fit the collected data, thereby reducing visual discrepancies during actual deployment and improving the success rate of task execution.

Moreover, due to the lightweight and agile design of the exoskeleton, we can conveniently place the power source and a compact computer in a backpack. As illustrated in Fig.~\ref{fig2}  (d), the exoskeleton system is worn for data collection in various scenarios. Considering the necessity for a third-person perspective camera, we can either place the camera at a fixed position in the external environment or mount the external camera on the chest to record the pose during data collection(similar to Dexcap\cite{dexcap2024}).

\begin{figure*}[thpb]
  \centering
  \includegraphics[width=1.0\linewidth]{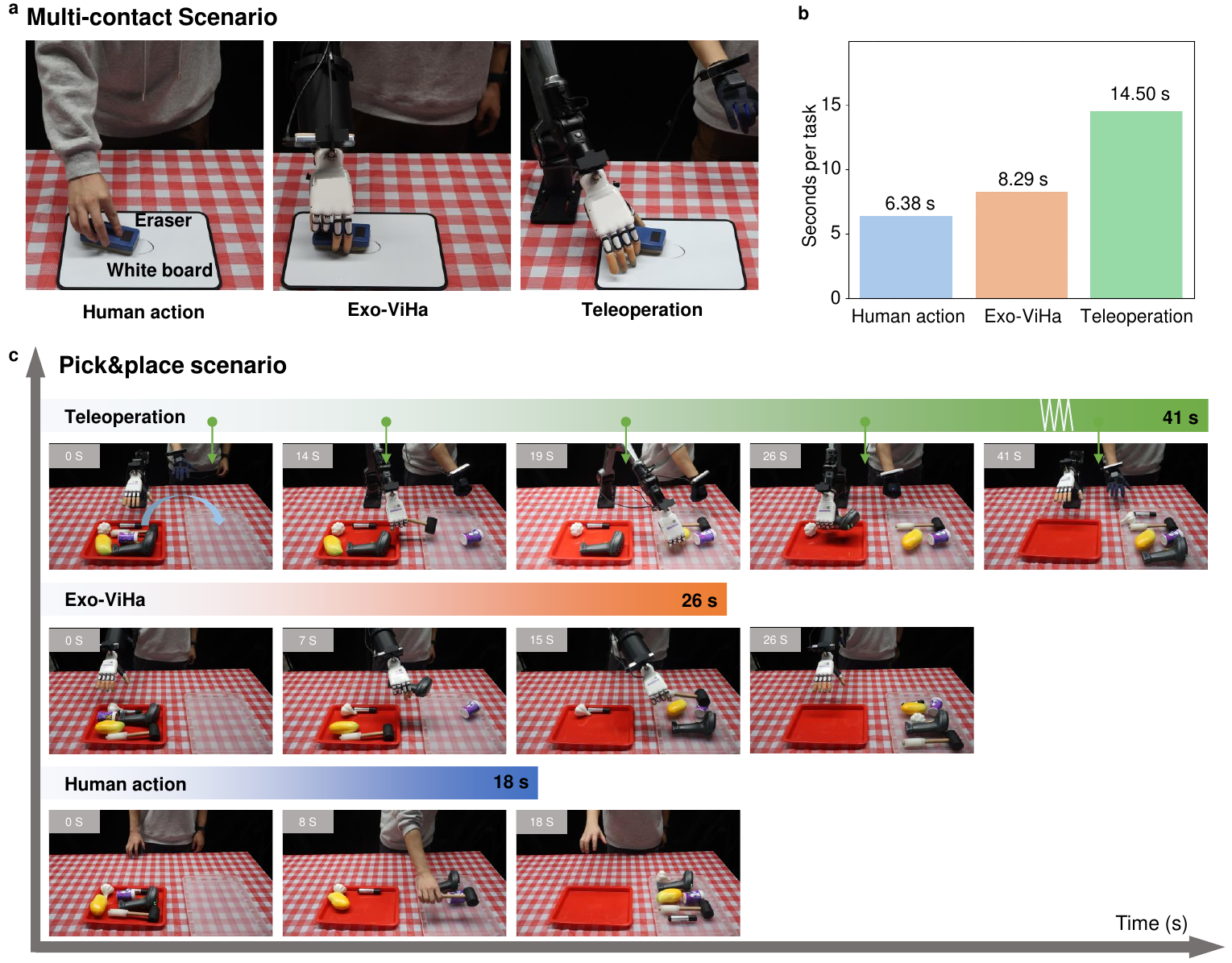}
  \caption{\textbf{Data collection efficiency comparision across different scenarios. }(a-b) Comparative analysis of whiteboard erasing tasks: (a) Multi-contact interaction scenarios demonstrating human action, Exo-ViHa and  teleoperation; (b) Quantitative time efficiency comparison across methods. (c) Pick and place scenario demonstrating the execution process for teleoperation, Exo-ViHa, and human action. 
}\label{fig5}
\end{figure*}
\subsection{Learning Framework}

Fig.~\ref{fig4} illustrates the overall learning framework of the proposed system.

During data collection using the exoskeleton, the user wears a motion capture glove to obtain real-time posture information of the hand beneath the arm exoskeleton 2. The data from the motion capture glove is then mapped onto the current dexterous hand, enabling it to replicate the user's hand posture. Simultaneously, real-time data from the dexterous hand is recorded to serve as input for the neural network model. After training, the model's output directly controls the dexterous hand, ensuring alignment between the input and output data formats. Although there may be some discrepancies in posture fitting between the motion capture glove and the dexterous hand, as long as the experimenter can control the dexterous hand to perform the corresponding tasks during data collection, the model's output will be sufficient to direct the dexterous hand in completing those tasks.

Meanwhile, we employ an Intel T265 SLAM camera to record the end-effector pose data of the exoskeleton, while calibration and measurement procedures are utilized to fit this data to the actual end-effector pose of the robotic arm:

\begin{equation} \mathbf{T}{\text{robot}} = \mathbf{T}{\text{calib}} \cdot \mathbf{T}_{\text{SLAM}} \end{equation}
where {T}{\text{calib}} represents the calibration matrix aligning the SLAM coordinate system with the target robotic arm's workspace.

Visual inputs comprise RGB streams from three perspectives: two third-view cameras and a wrist camera.

Finally, we chose to use the ACT model for training within the imitation learning framework. By utilizing the ACT model, we can ensure that the dexterous hand executes predetermined tasks efficiently and smoothly, thereby enhancing the task completion rate.

\section{Experiment}

\begin{figure*}[thpb]
  \centering
  \includegraphics[width=1.0\linewidth]{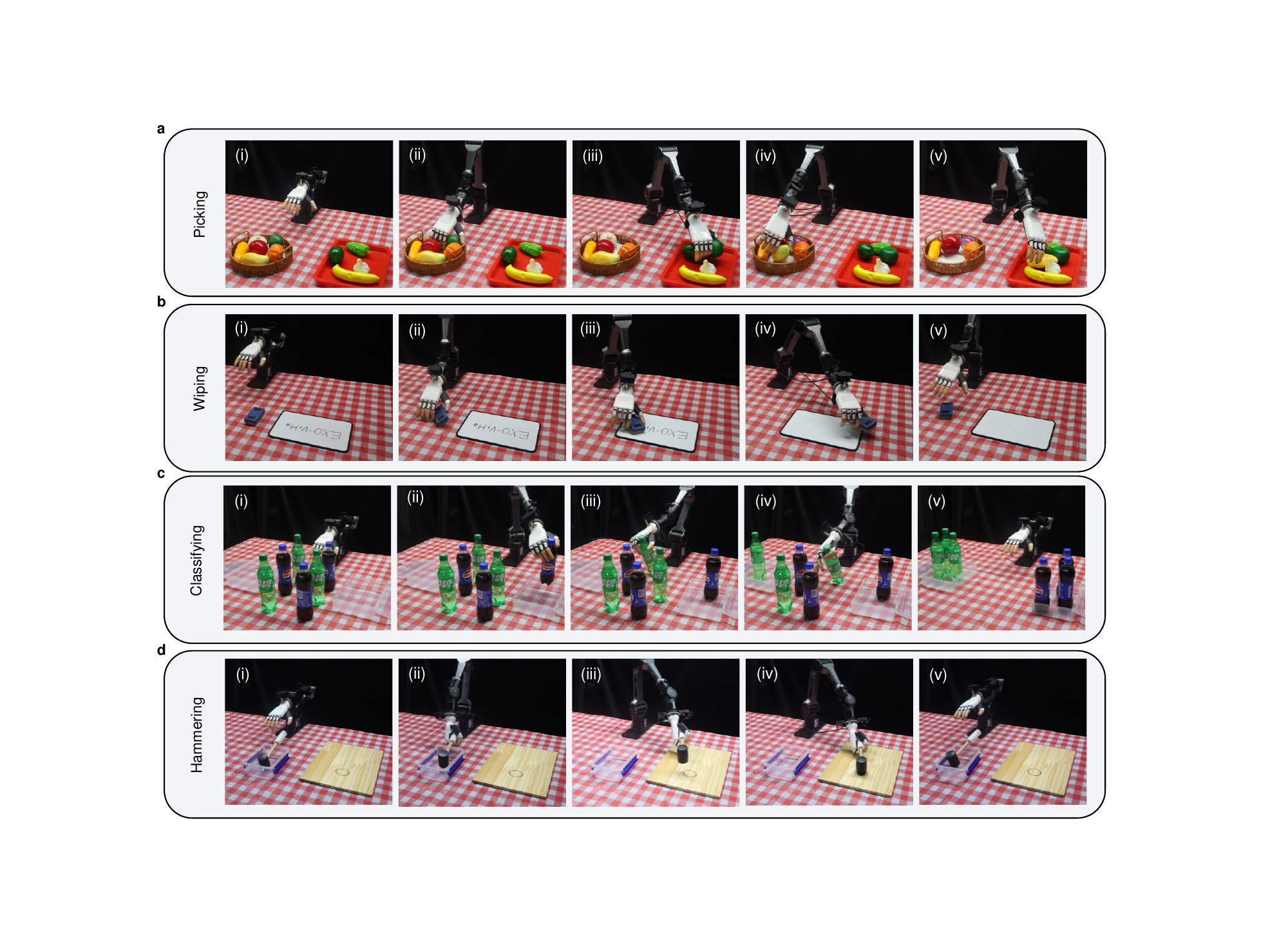}
  \caption{\textbf{Deployment of various robotic tasks.} (a) Picking: pick and place multiple objects of different shapes. (b) Wiping: hold the whiteboard wipe to wipe the whiteboard writing. (c) Classifying: sort the bottles based on their types. (d) Hammering: hold the hammer and hit the designated position.
}\label{fig6}
\end{figure*}

In this section, to evaluate the advantages of the proposed system, we conduct experiments on two key aspects of imitation learning: the efficiency of the data collection phase and the performance during real-world robot operation. The efficiency comparison of data collection using this system is shown in Fig.\ref{fig5} and Table \ref{tab:performance}, while the task examples of real robotic arm deployment after training with imitation learning are presented in Fig.\ref{fig6}.

\subsection{Collection Efficiency}

Currently, the primary form of data acquisition for imitative learning in dexterous manipulation tasks is represented by teleoperation. In this section, we compare the time required for different tasks using human hands versus traditional teleoperation with the timing of our system. 

Fig.~\ref{fig5} illustrates the time taken by the three different data acquisition methods to complete the task of wiping a whiteboard with an eraser. The results indicate that, for this multi-contact manipulation task, the completion time of our system is only a few seconds slower than that of human hands. In contrast, teleoperation, due to the lack of first-person perspective for observing object contact, often led to issues where the dexterous hand exerted excessive force, resulting in hard contact with the surface or gripping the eraser without making actual contact with the whiteboard. These factors contributed to an increased task completion time, as detailed in Fig.~\ref{fig5} (b).

Furthermore, when performing simpler tasks such as picking up and placing objects, our system demonstrates high efficiency. As shown in Fig.~\ref{fig5} (c), this task involved placing items of various sizes and shapes. It is evident that the time taken to complete the task using our system is only 8 seconds longer than that of autonomous human movement, while the use of teleoperation nearly doubled the time required.

\begin{table}[ht]
    \centering
    \caption{Performance metrics (average time and success rate) for robotic manipulation tasks during data collection and robot deployment using the Exo-ViHa system.}
    \begin{tabular}{|c|c|c|c|}
    \hline
    \multirow{2}{*}{Task} & \multicolumn{2}{c|}{\textbf{Data Collection}} & \multicolumn{1}{c|}{\textbf{Imitation}} \\
    \cline{2-3} \cline{4-4}
    & Avg. time (s) & Succ./Trials & Succ.Rate  \\ 
    \hline
    Pick-place objects     & 4.8 $\pm$ 0.9 & 29/30 & 86.6\% \\
    Sort six bottles       & 41.8 $\pm$ 7.8 & 27/30 & 76.6\% \\
    Hammer manipulation    & 12.4 $\pm$ 3.3 & 28/30 & 83.3\%\\
    Wipe whiteboard        & 12.9 $\pm$ 2.1 & 27/30 & 80.0\% \\
    \hline
    \end{tabular}
    \footnotesize
    \vspace{0.5em}

    Avg. time represent mean $\pm$ standard deviation (n=30 trials)
    \label{tab:performance}
\end{table}

    

Finally, we conducted tests on various task types for 30 iterations, as summarized in Table  \ref{tab:performance}. The system demonstrates reduced execution times and higher task completion rates across nearly all tasks

\subsection{Manipulation Performance}

For imitation learning, we conducted training on the LeRobot\cite{lerobot} platform using an NVIDIA RTX 4090 GPU. The training process consisted of a total of 160,000 offline steps. For each action, we conducted 30 episodes, with each episode completing the corresponding task. The batch size was set to 16, and the learning rate was configured to 2e-5.

In the Fig.~\ref{fig6},We deployed the Piper robotic arm along with the Inspire dexterous hand to accomplish the tasks depicted.  In Fig.~\ref{fig6} (a), The dexterous hand was used to grasp multiple fruit models and transfer them to the adjacent red plate. In Fig.~\ref{fig6} (b), Starting from the initial position, the dexterous hand picked up an eraser from the right side of the whiteboard and subsequently wiped the markings on the whiteboard from right to left. In Fig.~\ref{fig6} (c), On the table, there were six bottles of different colors; the dexterous hand sorted the bottles by color, placing each type into its corresponding plastic container. In Fig.~\ref{fig6} (d), The dexterous hand first grasped a hammer from the box, then lifted the hammer and moved it to a designated location, striking down into the target circle.

During the deployment phase, we conducted 30 trials for each task. As shown in Table II, the success rate for each task was generally around 80\%. The success rate was highest for the relatively simple grasping tasks, while the success rate for the classifying task was lower. This was due to the large size of the bottle, combined with the use of a full-grip method during data collection, which required precise positioning of the dexterous hand. A slight positional errors during deployment could cause the bottle to be knocked over. Additionally, in the Hammering and Wiping tasks, most failures occurred during the initial step of grasping the hammer and whiteboard eraser. In most cases, once the hammer and eraser were successfully grasped, the robotic arm was able to complete the subsequent tasks.

\section{Discussion and Future Work}
In this study, we introduced the Exo-ViHa system, a novel 3D printed exoskeleton designed to achieve a balance between data collection efficiency, consistency, and accuracy in dexterous tasks. The system enables users to have first-person visual feedback and direct tactile sensations during data collection. Additionally, the exoskeleton design ensures that the visual information captured by the cameras during both data collection and deployment phases is highly consistent, while the dexterous hand data is directly derived from the hand itself, significantly reducing discrepancies in visual and body alignment. Extensive experiments have demonstrated that the data collection efficiency achieved with this system is nearly identical to that of human actions, significantly surpassing the efficiency of traditional teleoperation methods. Furthermore, the success rate of automatic task deployment in multi-contact tasks after training reaches about 80\%, demonstrating the effectiveness and efficiency of the system.

Future work will focus on further structural improvements to the system: the forearm exoskeleton will be optimized to provide additional working space and comfort for the user's hand. Additionally, as data collection is a relatively lengthy process, improving the gravity compensation mechanism will be a key area of focus. Meanwhile, we will focus on evaluating the Exo-ViHa system in real-world settings, assessing its performance in collaborative tasks. Through extensive field studies, we aim to ensure the system's practicality and effectiveness across diverse applications.

\section*{Acknowledgements}

This work was supported by National Key R\&D Program of China (No.2024YFB3816000), Shenzhen Key Laboratory of Ubiquitous Data Enabling (No. ZDSYS20220527171406015), Guangdong Innovative and Entrepreneurial Research Team Program (2021ZT09L197), and Tsinghua Shenzhen International Graduate School-Shenzhen Pengrui Young Faculty Program of Shenzhen Pengrui Foundation (No. SZPR2023005).

\bibliographystyle{ieeetr}
\bibliography{main}
\end{document}